\documentclass[letterpaper, 10 pt, conference]{ieeeconf}  

\IEEEoverridecommandlockouts                              

\usepackage{amsmath,amsfonts}

\usepackage{algorithmic}
\usepackage{algorithm}
\usepackage{array}
\usepackage[caption=false,font=normalsize,labelfont=sf,textfont=sf]{subfig}
\usepackage{textcomp}
\usepackage{stfloats}
\usepackage{verbatim}
\usepackage{graphicx}
\usepackage{cite}
\usepackage{lipsum}
\usepackage[hyphens]{url}
\usepackage[hidelinks]{hyperref}
\usepackage{booktabs}
\usepackage{multirow}
\usepackage{gensymb}
\usepackage{soul,color}

\usepackage{soul}

\title{\LARGE \bf
F2BEV: Bird's Eye View Generation from Surround-View Fisheye Camera Images
for Automated Driving}

\author{Ekta U. Samani$^{1}$, Feng Tao$^{2}$, Harshavardhan R. Dasari$^{3}$, Sihao Ding$^{2}$, and Ashis G. Banerjee$^{4}$
\thanks{This work was supported in part by Volvo Car Technology USA LLC.}
\thanks{$^{1}$E. U. Samani is with the Department of Mechanical Engineering, University of Washington, Seattle, WA 98195, USA,
        {\tt\small ektas@uw.edu}}%
\thanks{$^{2}$F. Tao and S. Ding were with Volvo Cars, Sunnyvale, CA 94085, USA, during this work, 
        {\tt\small \{feng.tao, sihao.ding\}@volvocars.com}}%
\thanks{$^{3}$H. R. Dasari is with Volvo Cars, Sunnyvale, CA 94085, USA, 
        {\tt\small harshavardhan.reddy.dasari@volvocars.com}}%
\thanks{$^{4}$A. G. Banerjee is with the Department of Industrial \& Systems Engineering and the Department of Mechanical Engineering, University of Washington, Seattle, WA 98195, USA,
        {\tt\small ashisb@uw.edu}}%
}

\begin{document}

\maketitle
\thispagestyle{empty}
\pagestyle{empty}


\begin{abstract}

Bird's Eye View (BEV) representations are tremendously useful for perception-related automated driving tasks. However, generating BEVs from surround-view fisheye camera images is challenging due to the strong distortions introduced by such wide-angle lenses. We take the first step in addressing this challenge and introduce a baseline, F2BEV, to generate discretized BEV height maps and BEV semantic segmentation maps from fisheye images. F2BEV consists of a distortion-aware spatial cross attention module for querying and consolidating spatial information from fisheye image features in a transformer-style architecture followed by a task-specific head. We evaluate single-task and multi-task variants of F2BEV on our synthetic FB-SSEM dataset, all of which generate better BEV height and segmentation maps (in terms of the IoU) than a state-of-the-art BEV generation method operating on undistorted fisheye images. We also demonstrate discretized height map generation from real-world fisheye images using F2BEV. Our dataset is publicly available at \href{https://github.com/volvo-cars/FB-SSEM-dataset}{\url{https://github.com/volvo-cars/FB-SSEM-dataset}}
\end{abstract}



\section{Introduction}

Surround-view systems for automated driving typically use multiple wide-angle fisheye lens cameras to capture the near-field area around a car. Such cameras offer a larger field of view than pinhole cameras enabling complete coverage of a car's near-field area with fewer cameras. However, despite their ubiquity in the automotive industry, relatively few visual perception methods have been proposed for use in the case of fisheye camera images \cite{horgan2015vision, kumar2023surround}. 
Fisheye camera images exhibit strong radial distortion that cannot be corrected without reducing the field of view or introducing resampling artifacts \cite{kumar2020unrectdepthnet}. The spatially variant distortion also causes significant appearance variations for objects close to the cameras. As a result, monocular perception tasks, such as object detection and tracking, become more complex \cite{rashed2021generalized}. 
The complexity increases further in the multi-view setup of a surround view system as fusing predictions from the individual cameras requires complex post-processing steps \cite{fischer2022cc}. 
Such complexity can be avoided by performing perception tasks in a Bird's Eye View (BEV) representation generated from the multiple fisheye camera images.

BEV representations are rich in semantic information, such as the locations and scale of the objects in a car's surroundings, making them ideal for downstream real-world tasks. For instance, BEV height and segmentation maps can be useful in collision avoidance and motion planning. In addition, BEV maps provide a way to fuse information from different modalities and times. Generating BEV representations using surround-view cameras (which are placed parallel to the ground and facing outwards) is a challenging problem, even for pinhole cameras, as the transformations from the perspective view (PV) images to the BEV are ill-posed. Several efforts have been made 
to develop learning-based methods to generate BEV from PV images for complex real scenes \cite{ma2022vision}. However, generating BEV representations from fisheye images is still 
an open problem \cite{kumar2023surround}.

To the best of our knowledge, this work is the first reported attempt at obtaining BEV representations directly from surround-view fisheye camera images. Specifically, the contributions of our work are as follows:

\begin{itemize}
\item We introduce an effective baseline approach, F2BEV (i.e., Fisheye to BEV), incorporating fisheye projection and distortion models into a transformer-based BEV generation approach originally designed for PV images. We propose single-task and multi-task variants of F2BEV for individual and simultaneous generation of BEV segmentation and discretized height maps from fisheye images, respectively.

\item We present a synthetic dataset called FB-SSEM consisting of surround-view \underline{F}isheye camera images and \underline{B}EV maps from \underline{S}imulated \underline{S}equences of \underline{E}go car \underline{M}otion.

\item We report improved BEV segmentation and discretized height map generation from fisheye images using F2BEV over a state-of-the-art BEV generation method \cite{li2022bevformer} operating on undistorted fisheye images.

    
\end{itemize}


 \section{Related Work}

\subsection{Vision-based BEV Generation} 
BEV generation from camera images is often studied in the context of downstream perception tasks such as object detection/tracking, and lane/map segmentation. Recent literature considers BEV generation from a single PV (i.e., frontal view) or multiple PVs. In both the cases, the BEV generation methods either explicitly rely on geometry (i.e., the camera projection process) or implicitly learn the PV to BEV transformation using the camera geometry.

Projection relationships are explicitly incorporated in some methods using Inverse Perspective Mapping (IPM) \cite{mallot1991inverse} through preprocessing \cite{kim2019deep}, postprocessing \cite{sengupta2012automatic}, or feature transformation \cite{reiher2020sim2real}. IPM solves the ill-posed problem of mapping image pixels to points in the 3D space by assuming that all the points lie on a horizontal plane. Approaches based on generative adversarial networks 
have been explored to reduce the distortion resulting from this loss of height information \cite{zhu2018generative,bruls2019right}. Alternatively, depth prediction-based approaches have been proposed to preserve height information by lifting 2D feature maps to the 3D space \cite{philion2020lift, hu2021fiery, huang2021bevdet} or obtaining pseudo-LiDAR points \cite{you2020pseudo}.  

Another class of approaches implicitly incorporates camera geometry using a neural network to learn the mapping between a PV and the corresponding BEV. For example, the methods in \cite{lu2019monocular,pan2020cross,roddick2020predicting} employ a multi-layer perceptron to learn this transformation. Recently, transformer-based methods \cite{ma2022vision} have been introduced to boost the feature extraction process in the neural network and outperform the previous methods. In particular, the transformer-based methods perform the view transformation with the help of cross attention between the BEV queries and the input image features. 
For dense perception tasks (e.g., segmentation), dense query-based methods are used, where each query is associated with a predefined spatial location in the BEV space. Geometrical cues are often used in such methods to reduce memory complexity. For instance, camera calibration matrices have been used to obtain the reference points for the deformable attention modules \cite{li2022bevformer,lu2022learning} or to simplify global cross attention by pre-assigning vertical image scanlines to the BEV rays \cite{saha2022translating,gong2022gitnet,jiang2022polarformer}.

\subsection{Visual Perception Methods for Fisheye Images}
So far, relatively limited visual perception methods have been proposed for fisheye images owing to the distortion-related challenges. Standard vision pipelines (i.e., pipelines designed for pinhole camera images) cannot be applied to fisheye images without adaptations even for essential perception tasks such as object detection and semantic segmentation. For instance, the translational invariance leveraged in typical convolutional neural networks 
that achieve remarkable performance in imaging tasks is inapplicable for fisheye images due to the spatially variant distortion in them. Therefore, CNN-based approaches for semantic segmentation of fisheye images use specialized modules such as the overlapping pyramid pooling module \cite{deng2017cnn} or restricted deformable convolution \cite{deng2019restricted}. For geometry-based tasks, such as stereo matching and depth estimation, some approaches incorporate specific camera projection models into their framework \cite{hane2014real,won2019omnimvs} or use the camera parameters as a conditional input to their network \cite{kumar2021svdistnet}.

\section{Method}

\begin{figure*}[t]
    \centering
    \includegraphics[width=0.9\textwidth]{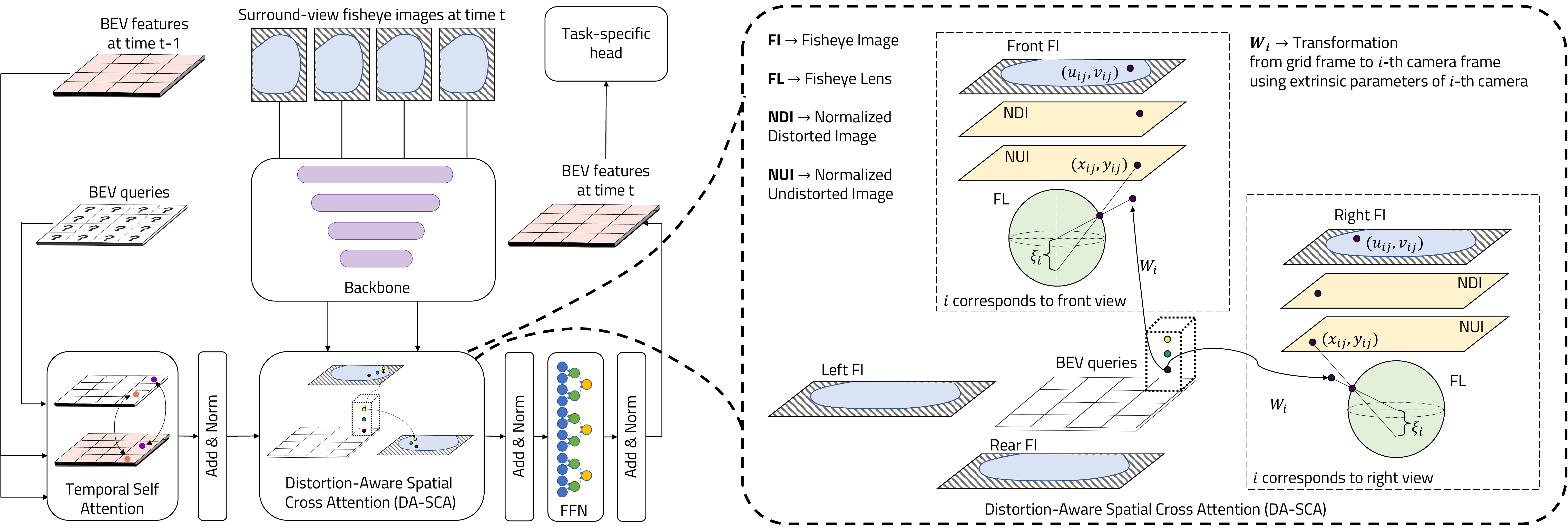}
        \caption{A schematic illustration of F2BEV for Bird's Eye View (BEV) generation from surround-view fisheye camera images} 
    \label{pipeline}
\end{figure*}

Given a set of surround-view fisheye images, our goal is to obtain a discretized BEV height map and a semantically segmented BEV map of the scene. To achieve this goal, we propose a dense query-based approach named F2BEV. The following subsections describe the main components of F2BEV, and Fig. \ref{pipeline} illustrates the overall architecture. 

\subsection{Fisheye Images to BEV Features}
\label{methoda}
First, we associate a query (with a learnable positional embedding), $q_p \in \mathbb{R}^d$, to every grid cell $p$ in a BEV plane of dimensions $h$ and $w$. The ego car is at the center of this BEV plane, and each grid cell represents a real-world cell of side $l$ meters. We use these queries in a temporal self attention module, as described in \cite{li2022bevformer}, to inquire about the temporal information from the BEV features at the previous timestep. Specifically, BEV features corresponding to the previous timestamp $t-1$ are aligned with the queries at the current timestamp $t$ using ego-motion and fed into the temporal self attention module.

Next, using a feature extraction backbone, we extract multi-scale feature maps from the images corresponding to the $n_\text{cam}$ surround-view fisheye cameras. Let $\mathcal{F}_t = \{ f_t^i, i \in \{1, \ldots, n_{\text{cam}}\}\}$ represent the extracted feature maps, where $f_t^i$ is the multi-scale feature map obtained from the the $i$-th view input fisheye image at time $t$. We then use a distortion-aware spatial cross attention module to obtain spatial information from these multi-scale features.

Our distortion-aware spatial cross-attention module is based on the camera model in \cite{mei2007single} and the spatial cross-attention in \cite{li2022bevformer}. First, we calculate the real-world location $(x_r, y_r)$ corresponding to every query $q_p$ located at $p = (x, y)$ as follows.

\begin{equation}
\begin{bmatrix}
    (x_r = (x - \frac{w}{2})l  & y_r = (y - \frac{h}{2})l 
\end{bmatrix}
\end{equation}

\noindent We then consider $n_a$ predefined heights to obtain a set of 3D reference points, $\mathcal{R}_p = \{ (x_r,y_r,z_j), j \in \{1, \ldots, n_a\} \}$, corresponding to the location $p$.

Similar to \cite{li2022bevformer}, we use the resource-efficient deformable attention \cite{zhu2020deformable} in our distortion-aware spatial cross attention and ensure that each query $q_p$ inquires about spatial information only in the relevant regions across all the camera views. Specifically, query $q_p$ only interacts with those camera views that have valid 2D projections corresponding to the points in $\mathcal{R}_p$. Note that the valid 2D projections are those whose coordinates correspond to the image region that is not covered by the camera or the ego car (i.e., the blue region of the fisheye images in Fig. \ref{pipeline}). We use $\mathcal{V}_{\text{valid}}$ to denote these views. The 2D projections of $\mathcal{R}_p$ in $\mathcal{V}_{\text{valid}}$ are then used as reference points of the query $q_p$ and the multi-scale feature maps corresponding to $\mathcal{V}_{\text{valid}}$ are sampled around them. A weighted sum of the sampled features is then computed, as follows, to obtain the distortion-aware spatial cross-attention (DA-SCA) output. 


\begin{equation}
    \text{DA-SCA}(q_p, \mathcal{F}_t) = \frac{1}{\mathcal{V}_{\text{valid}}} \sum_{i \in \mathcal{V}_{\text{valid}}} \sum_{j=1}^{n_a}\Delta(q_p, (u_{ij}, v_{ij}),f_t^i),
\end{equation}

\noindent where $\Delta$ represents deformable attention as described in \cite{zhu2020deformable}, and $(u_{ij},v_{ij})$ represent the pixel coordinates of the 2D point on the $i$-th camera view projected from the $j$-th 3D point in $\mathcal{R}_p$.

We use the camera model from \cite{mei2007single} to obtain $(u_{ij},v_{ij})$. It comprises a unified projection model approximating a large range of fisheye lenses and a radial-tangential distortion model. First, we transform the 3D reference points in $\mathcal{R}_p$ from the BEV grid frame to the $i$-th camera frame using the corresponding extrinsic parameters (rotation $R_i$ and translation $T_i$). The points in the camera frame are then projected onto a unit sphere and transformed into another reference frame via a translation of magnitude $\xi_i$ in the $z$ direction. Here $\xi_i$ represents the mirror parameter of the unified projection model for the $i$-th camera. These translated points are then projected onto the normalized undistorted image plane. Let $(x_{ij}, y_{ij})$ represent the coordinates of the $j$-th translated point in the normalized undistorted image plane (corresponding to the $i$-th camera). Last, we model the radial and tangential distortions using $[k_1,k_2]$ and $[k_3,k_4]$, respectively, and apply a projection matrix $K_i$ to obtain $(u_{ij},v_{ij})$ as follows.

\begin{equation}
\resizebox{0.48\textwidth}{!}{$
\begin{bmatrix}
u_{ij} \\
v_{ij} \\
1
\end{bmatrix}
= 
K_i\begin{bmatrix}
x_{ij} + \underbrace{x_{ij}\mathcal{L}(\rho_{ij})}_{\text{radial distortion}} + \underbrace{2k_3x_{ij}y_{ij} + k_4(\rho_{ij}^2 + 2x_{ij}^2)}_{\text{tangential distortion}}\\
y_{ij} +  \underbrace{y_{ij}\mathcal{L}(\rho_{ij})}_{\text{radial distortion}} + \underbrace{k_3(\rho^2 + 2y_{ij}^2) + 2k_4x_{ij}y_{ij}}_{\text{tangential distortion}}\\
1
\end{bmatrix}$}
\end{equation}

\noindent where $\rho_{ij} = \sqrt{x_{ij}^2 + y_{ij}^2}$, $\mathcal{L}(\rho) = (k_1 \rho^2 + k_2\rho^4)$, and  $K_i$ is obtained from Eq. (\ref{k}).

\begin{equation}
K_i = \begin{bmatrix}
\gamma_{1i} & \alpha_i \gamma_{1i} & c_{1i} \\
 0 & \gamma_{2i} & c_2i \\
0 & 0 & 1
\end{bmatrix},    
\label{k}
\end{equation}

\noindent where $\alpha_i$ is the skew, $[\gamma_{1i},\gamma_{2i}]$ are the focal lengths, and $[c_{1i},c_{2i}]$ are the principal point coordinates of the $i$-th camera.

Following the standard transformer architecture, the BEV queries are then fed to a feedforward network to obtain the BEV features at time $t$. The features are then passed to task-specific heads to compute the BEV maps.

\subsection{BEV features to BEV maps}
Different 
task-specific heads can be designed to obtain the BEV maps from the BEV features. We consider 
two categories of 
heads: single-task heads and multi-task heads.

Two types of single-task heads are considered to obtain the BEV maps. The first type is an attention-based head, which follows the mask decoder in \cite{li2022panoptic}. It uses one learnable query to represent every height class (or semantic class in the case of segmentation map generation). The final BEV maps are generated based on the attention maps from the multi-head attention modules using deep supervision. Deep supervision of attention masks from every layer allows the multi-head attention modules to focus on meaningful semantic regions. The second type of single-task head is a convolution-based single-task head. It consists of multiple cascaded upsampling blocks followed by a prediction block with a $3 \times 3$ convolution layer and an activation layer. Each block consists of a $2\times$ upsampling (using bilinear interpolation) operation, a dropout layer, a $3 \times 3$ convolution layer, and a ReLU activation layer successively. The upsampling blocks enable the generation of higher-resolution BEV maps without increasing the number of queries (i.e., reducing $l$), as opposed to the attention-based head that generates the BEV maps at the same resolution as the BEV features.

BEV height map generation and segmentation map generation tasks are inherently related. Moreover, downstream perception tasks in an automated driving system benefit from both the types of maps and their simultaneous generation is more computationally efficient than generating them separately. Therefore, we extend the single-task heads to obtain attention-based and convolution-based multi-task heads. In the attention-based multi-task head, we use one learnable query per height class and one learnable query per semantic class. Queries associated with a height class are fed to multi-head attention modules deeply supervised using the corresponding ground truth. Similarly, queries associated with a semantic class are fed to multi-head attention modules deeply supervised using ground truth segmentation maps. As with the single-task head, final BEV maps are generated from the corresponding attention maps. On the other hand, the convolution-based multi-task head consists of two branches, one for each task. Both branches share cascaded upsampling blocks before splitting into their respective prediction blocks.

\section{FB-SSEM Dataset}
A lack of suitable datasets is one of the reasons for the relatively sparse literature on perception methods for fisheye images. Augmentation techniques such as 7-DoF \cite{ye2020universal} and zoom augmentation \cite{deng2017cnn, deng2019restricted} have been proposed to convert perspective images into fisheye images. Recently, large-scale synthetic \cite{sekkat2020omniscape, sekkat2022synwoodscape} and real-world datasets \cite{maddern20171,yogamani2019woodscape,liao2022kitti} have been proposed for semantic segmentation and object detection tasks. None of these datasets, except SynWoodScape, consist of ground truth BEV height maps and segmentations. Each sample in SynWoodScape represents a random spawning of the ego vehicle and is associated with only one previous timestep. It does not capture temporal information related to the sequential motion of the ego car that has been shown to help BEV generation \cite{li2022bevformer}.
Therefore, we propose a new synthetic dataset, FB-SSEM consisting of surround-view fisheye camera images and BEV maps generated from the sequential motion of an ego car along with the corresponding ego-motion information.

We use the Unity game engine \cite{technologies} to simulate a parking lot environment for our dataset. The parking lot consists of parked cars/trucks, buses, electric vehicle (EV) charging stations of varying dimensions, and large containers of varying heights (on the boundaries). All the vehicles in the parking lot, except the ego car, are static. For the ego car, we use a forward-looking wide camera to simulate its four surround-view fisheye cameras \cite{rashed2021generalized}. Our dataset consists of 20 sequences of daytime ego car motion (at a speed of $0.35$ m/s) through the parking lot. Each sequence represents a different parking lot setup, i.e., different placement of all the vehicles in the lot and ground textures. Each sequence consists of 1000 samples recorded at a frequency of $2$ Hz; each sample consists of RGB images from the four car-mounted fisheye cameras (i.e., front, left, rear, and right cameras) and the BEV camera. Corresponding semantic segmentation maps for all five views and normalized height maps for the BEV are also generated. In addition, ego-motion information (3D rotation and translation) corresponding to every sample is obtained. We consider five classes for the BEV segmentation map: car (ego car and parked cars/trucks), bus, EV charger, ground, and a non-driveable area. For our work, we discretize the height maps 
into three primary classes: above car level, below car level, and at car level.


\section{Experiments}

\subsection{Implementation details}
We consider four cameras in our surround view system to generate a BEV plane of dimensions $16.67$m $\times16.67$m. We choose $l = 0.33$m and associate $2,500$ queries with the BEV plane. We use PyTorch and follow the implementation of temporal self attention from \cite{li2022bevformer} and attention-based task head from \cite{li2022panoptic}. For the distortion-aware spatial cross attention, we consider three height anchors to obtain $R_p$ and set them at $0$m, $0.25$m, and $1.8$m above the ground, respectively. The intrinsic and extrinsic camera parameters used in the distortion-aware spatial cross attention are obtained using the OpenCV's omnidir camera calibration module. We refer the reader to Section \ref{designchoices} for a discussion on the above choices.

Feature extraction from the fisheye images is performed using a ResNet34 with a feature pyramid network for attention-based task heads and a bidirectional feature pyramid network when using convolution-based task heads. The transformer block consisting of the temporal self-attention, the distortion-aware spatial cross attention, and the feedforward network is repeated three times to obtain refined BEV features. Similarly, the attention-based task heads have three multi-head attention modules per task. The convolution-based heads consist of three copies of upsampling blocks.

We an Ubuntu 20.04 
AWS EC2 instance equipped with an NVIDIA A10G Tensor Core GPU for training and testing. We evaluate the performance of single-task and multi-task F2BEV variants and compare it with the state-of-the-art BEVFormer \cite{li2022bevformer}. We undistort the fisheye images before feeding them to BEVFormer because it expects pinhole camera images as input. Focal lengths for undistortion are selected based on OpenCV's recommendation for perspective rectification \cite{opencv}. We divide every sequence using a (70-15-15)$\%$ train-val-test split and train models first using categorical cross-entropy loss, followed by tuning with multi-class focal loss. We consistently use the random sampling strategy used in BEVFormer to augment ego-motion diversity. During testing, we sequentially obtain predictions for each sample; i.e., we save BEV features from the previous timestamp and use them for prediction at the current timestamp.

\subsection{Results}
\subsubsection{\textbf{Results on the simulated dataset}}
Table \ref{heightresults} and Table \ref{segresults} show the discretized height map and segmentation map generation performance, respectively. F2BEV outperforms BEVFormer (which uses undistorted images as input) in both the tasks regardless of the type of task-specific head. 

\begin{table}[]
\centering
\caption{Performance comparison of F2BEV variants with BEVFormer (using undistorted images) for discretized BEV height map generation }
\label{heightresults}
\resizebox{\columnwidth}{!}{
\begin{tabular}{@{}c|ccc|cc@{}}
\toprule
\multirow{3}{*}{Method} & \multicolumn{3}{c}{Class-wise performance   (IoU)}                          & \multicolumn{2}{|c}{Overall performance}           \\ \cmidrule(l){2-6} 
                          & Below car & At car          & Above car & Mean  & Freq. Wt. \\
                          & level     & level           & Level     &  IoU         & IoU\text{*}       \\ \midrule 
F2BEV-Attn-ST            & 0.9786         & 0.8470                  & 0.8976         & 0.9078         & 0.8640         \\
F2BEV-Conv-ST             & 0.9732    & \textbf{0.8496} & 0.8459    & 0.8896   & 0.8526    \\
F2BEV-Attn-MT            & \textbf{0.9787}         & 0.8485                  & \textbf{0.9044}         & \textbf{0.9106}         & \textbf{0.8673}         \\
F2BEV-Conv-MT           & 0.9743    & 0.8478          & 0.8617    & 0.8946   & 0.8484    \\
Undistort $\rightarrow$   BEVFormer  & 0.9629 & 0.7656 & 0.8553 & 0.8612 & 0.7957 \\
\bottomrule 
\end{tabular}%
}
\end{table}

\begin{table*}[]
\centering
\caption{Performance comparison of F2BEV variants with BEVFormer (using undistorted images) for BEV segmentation map generation}
\label{segresults}
\resizebox{\textwidth}{!}{%
\begin{tabular}{@{}c|ccccc|cc@{}}
\toprule
\multirow{2}{*}{Method}           & \multicolumn{5}{c}{Class-wise performance   (IoU)}                             & \multicolumn{2}{|c}{Overall performance} \\ \cmidrule(l){2-8} 
                & Ground & EV charger & Non-driveable area & Bus             & Car    & Mean IoU & Freq. Wt. IoU\text{*}   \\ \midrule 
F2BEV-Attn-ST                     & \textbf{0.9779} & \textbf{0.6971} & 0.7293          & 0.8482 & 0.8987 & \textbf{0.8303}         & 0.8587        \\
F2BEV-Conv-ST   & 0.9734 & 0.6399     & 0.6851             & 0.8876          & 0.8848 & 0.8142   & \textbf{0.8624} \\
F2BEV-Attn-MT & 0.9776 & 0.6244     & 0.7597             & 0.8531            & \textbf{0.9011} & 0.8232   & 0.8591          \\
F2BEV-Conv-MT & 0.9743 & 0.6308     & 0.7105             & \textbf{0.9066} & 0.8847 & 0.8214   & 0.8560          \\
Undistort $\rightarrow$ BEVFormer & 0.9679          & 0.6385          & \textbf{0.8222} & 0.8015 & 0.8408          & 0.8142                  & 0.8149        \\ \bottomrule
\end{tabular}%
}
\end{table*}


We observe from Table \ref{heightresults} that both single-task and multi-task attention-based variants\footnote{Since attention-based variants generate predictions of a smaller resolution (but same physical area), we resize the ground truth to compute IoU.} (i.e., F2BEV-Attn-ST and F2BEV-Attn-MT, respectively) achieve slightly better overall performance than convolution-based single-task and multi-task variants (i.e., F2BEV-Conv-ST and F2BEV-Conv-MT, respectively) in the case of discretized height map generation. Moreover, F2BEV-Attn-MT achieves the best overall performance, with approximately $9\%$ improvement over BEVFormer in terms of the frequency-weighted IoU metric\footnote{The frequency-weighted IoUs are obtained by ignoring the background class (i.e., the below car level class and the ground class)}. It is also worth noting that both the multi-task variants achieve an overall performance comparable to their single-task counterparts while simultaneously generating BEV segmentation maps. We discuss the segmentation performance of the multi-task variants later in this subsection.

\begin{figure*}[t]
    \centering
    \includegraphics[width=0.89\textwidth]{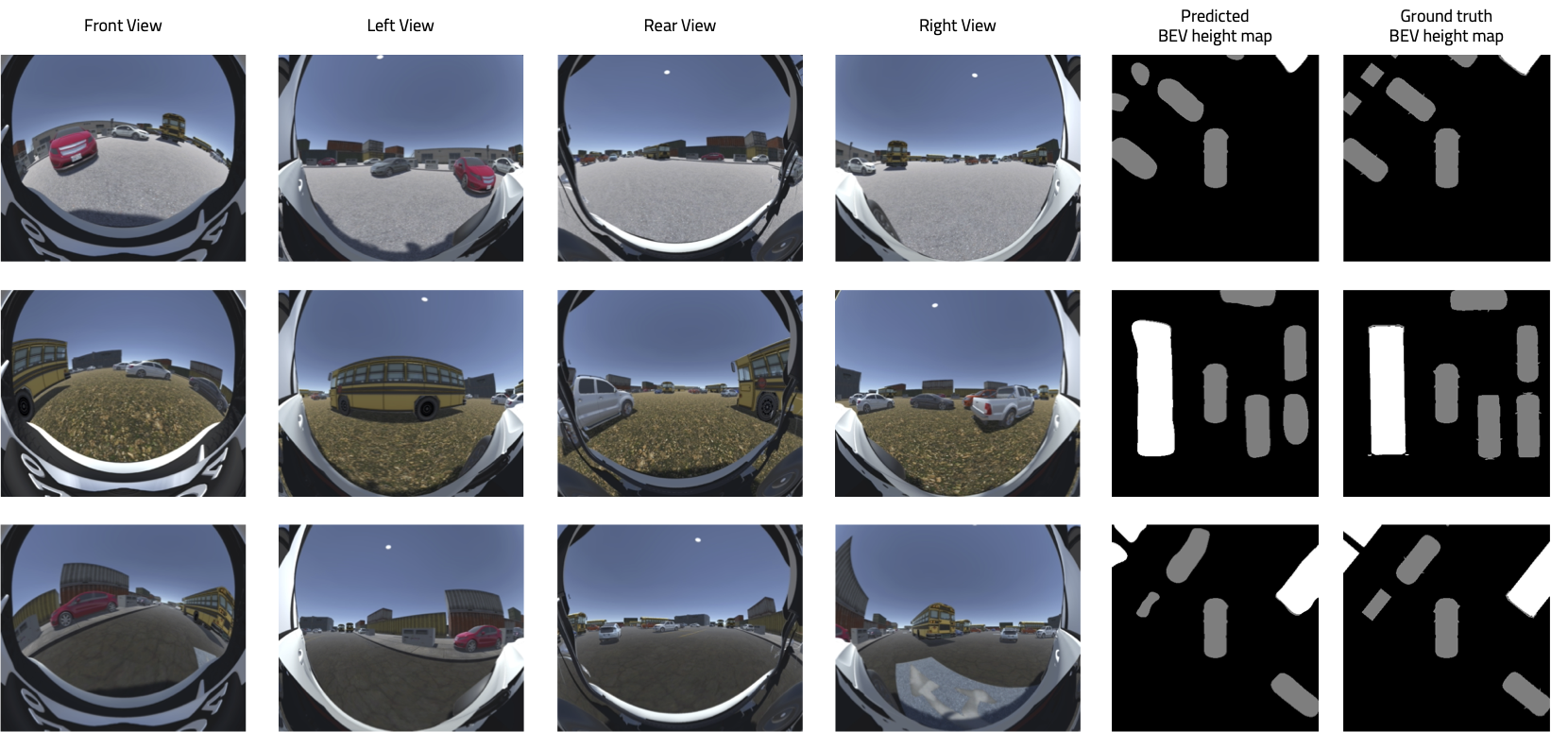}
        \caption{Samples of discretized BEV height maps obtained from simulated fisheye images using F2BEV-Conv-ST. In the BEVs, black color represents the 'below car level' class, gray represents the 'at car level' class, and white represents the 'above car level' class.}

    \label{htimages}
\end{figure*}

Fig. \ref{htimages} shows some discretized height map generation results. Note that we consistently include the relatively higher resolution results obtained using the F2BEV-Conv-ST variant in all the figures for better visualization. The first and second rows show F2BEV's ability to consolidate information from the different cameras. Specifically, F2BEV correctly places objects in the BEV even when they are entirely occluded in one camera view and partially visible in another (e.g., the EV charger behind the red car in the first row). We also see correct placements of long objects that are partially visible in one view and completely visible in another (e.g., the bus in the second row). The method also demonstrates an implicit understanding of the relative distances of the objects from the ego car by ignoring all the relatively distant objects present in the fisheye images during BEV generation. The results further illustrate the benefit of our distortion-aware approach in inferring and preserving the height information during BEV generation. For instance, the first and second rows show the correct height classification for vehicles strongly distorted due to their proximity to the ego car. 

\begin{figure*}[t]
    \centering
    \includegraphics[width=0.89\textwidth]{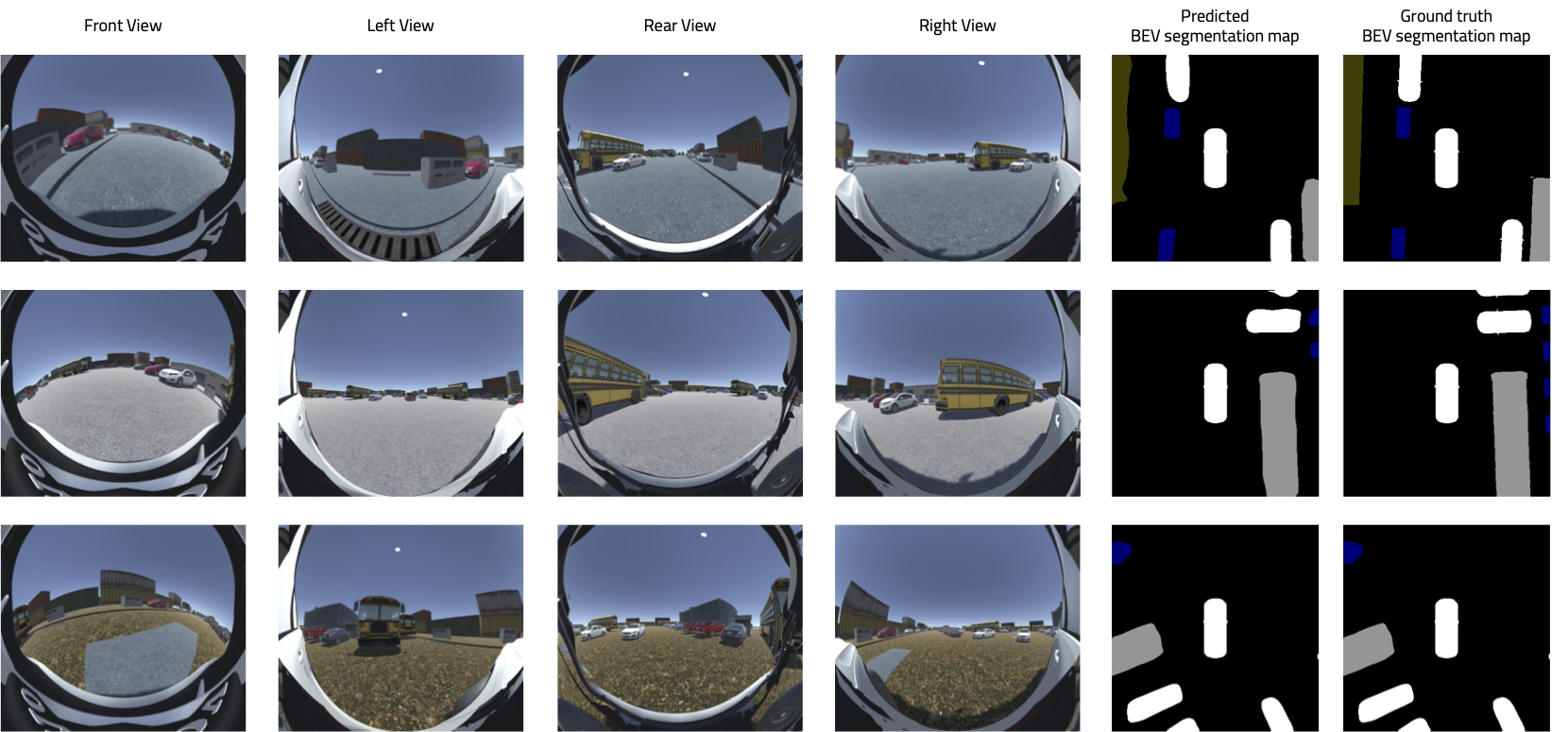}
        \caption{Sample BEV segmentation map predictions obtained from simulated fisheye images using F2BEV-Conv-ST. In the BEVs, black color denotes the ground class, white denotes the car class, gray denotes the bus class, blue denotes the EV charger class, and olive green denotes the non-driveable area.}

    \label{segimages}
\end{figure*}

In the case of segmentation map generation with five semantic classes, we note From Table \ref{segresults} that single-task variants of F2BEV achieve slightly better overall performance than their multi-task counterparts. Moreover, F2BEV-Conv-ST achieves the best overall performance, with a $5.8\%$ improvement over BEVFormer in terms of the frequency-weighted IoU metric. As in the case of discretized height map generation, all the F2BEV variants outperform BEVFormer operating on undistorted images. We also see successful consolidation of information from multiple cameras in the generated BEV segmentation maps. For instance, Fig. \ref{segimages} shows the correct placement of the EV charger behind the white car in the second row and the bus in the third row. We refer the reader to Section \ref{segperformancediscussion} for further discussion on class-wise segmentation performance, particularly the EV charger and non-driveable area classes.


\subsubsection{\textbf{Results on real-world images}}
\label{realworldresults}
\begin{figure*}[t]
    \centering
    \includegraphics[width=0.84\textwidth]{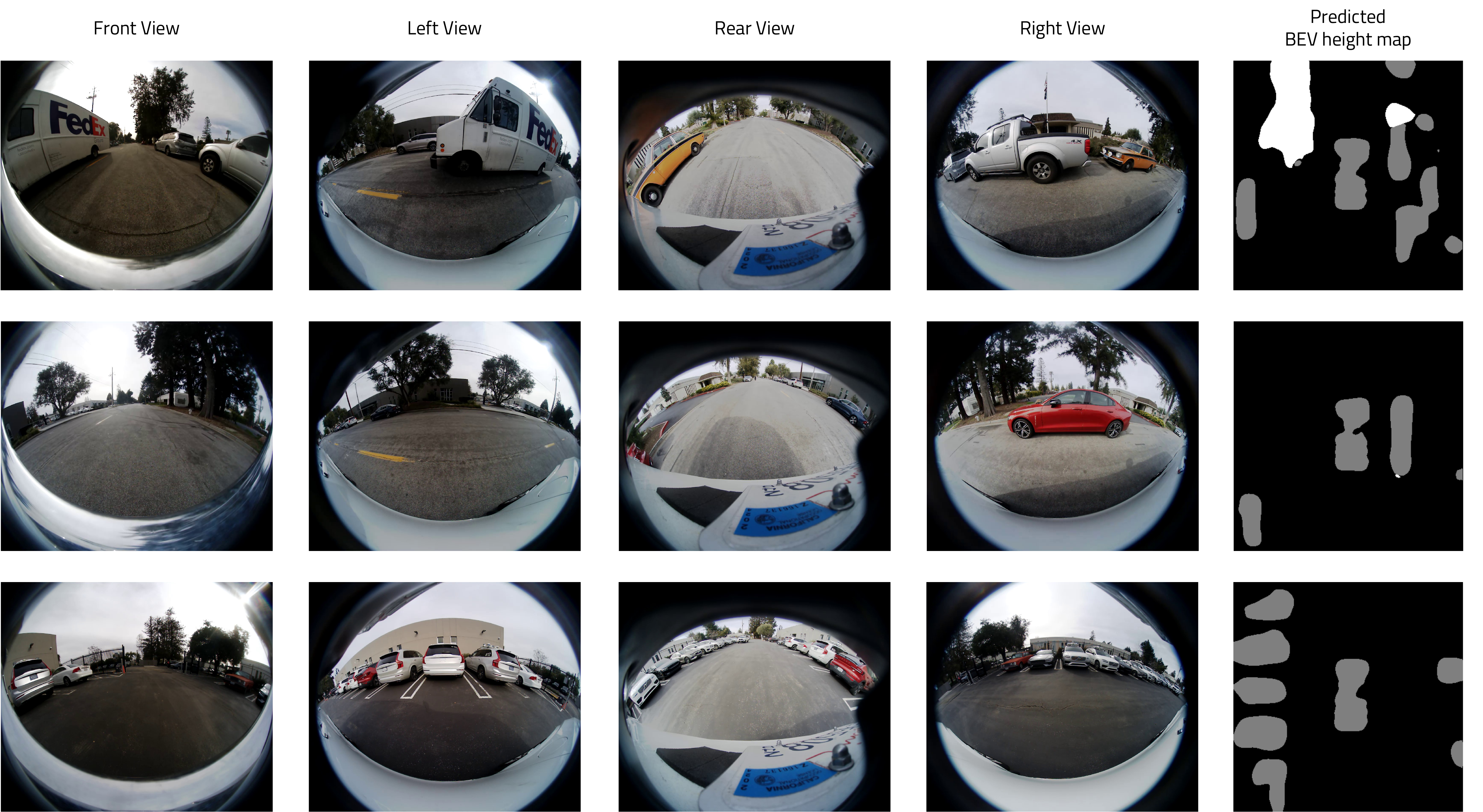}
        \caption{Samples of discretized BEV height maps obtained from real-world fisheye images using F2BEV-Conv-ST. In the BEVs, black color represents the 'below car level' class, gray represents the 'at car level' class, and white represents the 'above car level' class.}

    \label{realimages}
\end{figure*}

We also test the performance of F2BEV on real-world fisheye images captured using four $210\degree$ field of view cameras mounted on a Volvo XC90 car. We record ego-motion information using an IMU onboard the ego car with real-time noise filtering. Due to the unavailability of ground truth in this case, we only report a qualitative performance evaluation. Fig. \ref{realimages} shows a few sample real-world images.

The gap between the simulated samples in Fig. \ref{htimages} and the real-world samples in Fig. \ref{realimages} is considerably large. To reduce this gap, we train an F2BEV-Conv-ST model for discretized height map generation using simulated data styled as real-world data. We run neural style transfer \cite{gatys2015neural} on the simulated FB-SSEM dataset images using samples of real-world images as style images.
Additionally, we note that the simulated fisheye cameras differ from those mounted on the actual car in terms of the valid regions of the image (i.e., regions not covered by the camera or the ego car). We account for this difference during training by predefining the valid region for every camera in the training data to be the same as that of the corresponding real-world camera. We then use the trained model for predicting the discretized BEV height maps for the real-world data. 

Fig. \ref{realimages} shows the generated height maps for a few sample images\footnote{We only report height map generation performance because the output classes considered during training and testing are the same, unlike for segmentation map generation.}. The first and second row show instances of the car's motion on a street. We see the successful placement of a delivery truck and street-parked cars in the corresponding height maps. In the third row, we see an instance of the car's motion in a filled parking lot. We note from the fisheye images that the ego car is closer to the cars parked on its left than most of the cars on its right, which is also reflected in the 
height map. A video of the performance on real-world and simulated images is included in the supplementary materials. 




\subsection{Discussion}
\subsubsection{\textbf{Design choices}}
\label{designchoices}
We use 2,500 queries associated with the BEV plane in all our experiments. This choice directly results from the memory limitations of using a single GPU for training. A higher number of queries can be used in the case of multi-GPU training. A higher number of queries can help generate higher-resolution feature maps using the attention-based heads if $l$ is reduced. It can also help 
segment objects that appear narrow in the BEV maps, such as street lights and lane markings. Alternatively, increasing the number of queries without changing $l$ can help generate BEV for a physically larger area. In our experiments, every query $q_p$ is associated with three height anchors to obtain $\mathcal{R}_p$. These height anchors are set based on the approximate ground clearance and the height of the ego car. A higher number of height anchors can also be considered to obtain discretized height maps with more than three height classes. 


Our distortion-aware spatial cross attention module uses unified projection that approximates fisheye lenses and performs distortion correction based on a radial tangential distortion model, as in \cite{mei2007single}, to obtain 2D reference points from $\mathcal{R}_p$. Our results validate the suitability of incorporating distortion within the network to obtain BEV maps 
instead of undistorting the imagess themselves. We believe distortion-aware attention modules based on other projection models \cite{caruso2015large,khomutenko2015enhanced, usenko2018double, yogamani2019woodscape} can also be incorporated into standard BEV generation methods designed for PV images to achieve better performance than image undistortion-based approaches. 


\subsubsection{\textbf{Segmentation of the EV charger and non-driveable area classes}}
\label{segperformancediscussion}
In the second row of Fig. \ref{segimages}, we note that two EV chargers are missing in the predicted BEV map. This observation is consistent with our expectations; the two EV chargers do not appear in the corresponding camera views as they completely occluded by the bus. The ground truth BEV does not capture such occlusions because it is generated using a perspective camera above the ground. Our current IoU-based quantitative evaluation does not account for such scenarios, which is one of the reasons for the relatively lower segmentation performance for the EV charger class. 

In the case of the non-driveable area, BEVFormer achieves higher IoU as compared to F2BEV. We attribute this trend to the relatively sparse sampling of fisheye image features compared to the undistorted image features. Specifically, certain sections of the non-driveable area are much taller than our tallest height anchor, which is 1.8 m (approximate car height). Therefore, the input image features are thinly sampled in the region corresponding to the non-driveable area. The lack of sampling affects F2BEV 
more than BEVFormer because the non-driveable area occupies fewer pixels in the fisheye images than in the undistorted images (due to the difference in their fields of view). We believe this can be addressed by increasing the height anchors in a manner representative of the semantic classes in the environment.


\subsubsection{\textbf{Distortion-aware approach versus input image undistortion}}
We demonstrate in Tables \ref{heightresults} and \ref{segresults} that using distortion-aware spatial cross attention in F2BEV achieves better performance than using undistorted images with BEVFormer, even in our dataset's relatively simple parking lot environment. Moreover, the undistortion of a fisheye image into a perspective image results in a loss of information due to the reduced field of view. Such a loss of information would lead to performance drops when generating the BEV maps covering a larger area. A more involved undistortion method to obtain multiple undistorted images from a single fisheye image can also make consolidating information into a single BEV using cross-attention modules more challenging.



\addtolength{\textheight}{-0.5cm}
\section{CONCLUSIONS}

In this work, we introduce F2BEV: a first-of-its-kind approach for generating BEV maps directly from surround-view fisheye camera images. It incorporates fisheye projection and distortion models into a spatial cross attention module to query spatial information in fisheye image features and consolidate it into a BEV representation. Following a state-of-the-art BEV generation method for pinhole camera images, the distortion-aware spatial cross attention module is placed in a transformer-style architecture with a temporal self attention module and feedforward network to obtain BEV features. We consider single-task and multi-task heads to generate discretized height maps and semantic segmentation maps from the generated BEV features. On our synthetic FB-SSEM dataset, F2BEV outperforms (in terms of the IoU) a state-of-the-art BEV generation method \cite{li2022bevformer} operating on undistorted fisheye images, regardless of the choice of task heads. We also demonstrate F2BEV's performance in generating the BEV maps from real-world fisheye images obtained from an actual car. In addition, our FB-SSEM dataset is publicly available to encourage further research in this domain, such as developing distortion-aware modules that are independent of camera calibration parameters \cite{peng2023bevsegformer} to obtain different types of BEV maps (e.g., continuous height maps, image-like RGB maps) for automated driving tasks.






\section*{ACKNOWLEDGMENT}
We acknowledge the contributions of Zhipeng Liu, John Nousiainen, Chihiro Suga, and Jeremy White in building the simulated environment for generating the FB-SSEM dataset.




\bibliography{references}

\bibliographystyle{IEEEtran}

\end{document}